\pdfoutput=1
\documentclass[letterpaper]{article}
\usepackage{aaai2027}
\nocopyright  
\makeatletter
\def\@fnsymbol#1{\ensuremath{\ifcase#1\or *\or \dagger\or \ddagger\or \mathsection\or \mathparagraph\else\@ctrerr\fi}}
\makeatother
\usepackage[hyphens]{url}  
\usepackage{graphicx} 
\usepackage{natbib}  
\usepackage{caption} 
\usepackage{algorithm}
\usepackage{algorithmic}

\usepackage{newfloat}
\usepackage{listings}
\DeclareCaptionStyle{ruled}{labelfont=normalfont,labelsep=colon,strut=off} 
\floatstyle{ruled}
\newfloat{listing}{tb}{lst}{}
\floatname{listing}{Listing}

\usepackage{booktabs}
\usepackage{amsmath,amssymb}

\usepackage{multirow}
\usepackage{colortbl}
\definecolor{gamebg}{gray}{0.90}     %
\definecolor{gameplusbg}{gray}{0.80} %
\definecolor{basegray}{gray}{0.5}    %

\title{CRAFT: Compression via Recursive Adaptive Fusion of Video Tokens for Vision-Language Models}

\author{
    Yu Chen\textsuperscript{\rm 1,2}\equalcontrib,
    Xiaohong Li\textsuperscript{\rm 1,2}\equalcontrib,
    Xiaole Wang\textsuperscript{\rm 1,2,3}\equalcontrib,
    Jianjin Zhang\textsuperscript{\rm 1,2},
    Jun Sun\textsuperscript{\rm 3},
    Yafeng Deng\textsuperscript{\rm 1,2}\corresponding
}
\affiliations{
    \textsuperscript{\rm 1}Evermind\quad
    \textsuperscript{\rm 2}Shanda Group\quad
    \textsuperscript{\rm 3}Peking University
}

\begin{document}

\maketitle

\begin{abstract}
In video understanding, vision-language models (VLMs) must ingest massive numbers of visual tokens, causing the computational and memory cost of the prefill stage to rise sharply. Such visual sequences are highly redundant along the spatio-temporal dimension, yet a high compression ratio is often accompanied by the loss of critical details.
Existing token-compression methods either employ heuristic, training-free compression with limited content adaptivity or introduce additional modules that require expensive alignment training, leaving the trade-off between efficiency and adaptivity unresolved.
To alleviate this limitation, we propose \textbf{CRAFT}: \textbf{C}ompression via \textbf{R}ecursive \textbf{A}daptive \textbf{F}usion of Video \textbf{T}okens.
CRAFT recursively merges tokens by decoupling parameter-free token selection from learnable token fusion: global similarity determines which tokens to merge, while a position-aware weighting module and a content-adaptive channel-wise gate learn how to fuse them. The whole compression pipeline is query-agnostic.
Because every retained token is a linear combination of the original tokens, CRAFT preserves their true spatio-temporal coordinates and stays aligned with the pre-trained language model's input distribution. Experiments on multiple representative video benchmarks show that CRAFT consistently outperforms prior state-of-the-art token-compression methods. At about $8\times$ compression, it retains roughly $97\%$ of the backbone's average accuracy and shows significant efficiency improvement. 
\end{abstract}

\begin{figure}[t]
    \centering
    \includegraphics[width=\columnwidth]{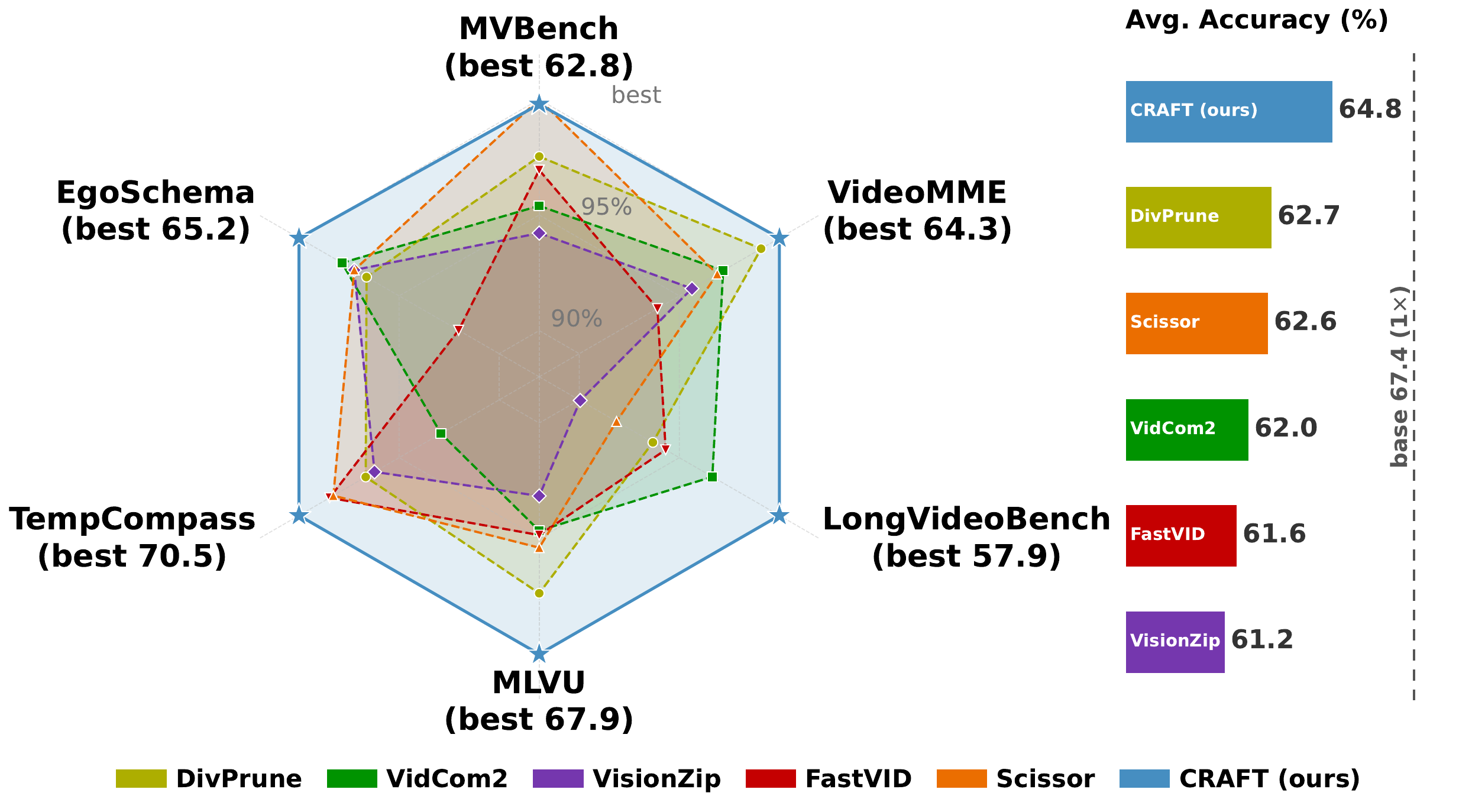}
    \caption{\textbf{Left:} six-benchmark accuracy comparison at a fixed $8\times$ compression ratio, normalized per benchmark to the best score among the six methods shown (outer vertex $=$ best). \textbf{Right:} benchmark-averaged accuracy; the dashed line marks the uncompressed backbone ($67.4$).}
    \label{fig:teaser}
\end{figure}

\section{Introduction}

Video understanding has become an important frontier direction for multimodal large language models (MLLMs)~\citep{bai2025qwen25vl, chen2025longvila, shu2025videoxl}. However, as visual tokens grow with the product of video frame count and spatial resolution, reaching tens of thousands of tokens for a single video, the prefill self-attention cost, which grows quadratically, and the decoding-stage KV-cache memory footprint, which accumulates linearly, become serious bottlenecks~\citep{shao2025survey, kong2025beyond}. Therefore, substantially shrinking the scale of visual tokens while preserving semantics is a straightforward path to improving video reasoning efficiency. Masked autoencoders~\citep{he2022mae, tong2022videomae} and attention analysis~\citep{chen2024fastv} confirm that visual sequences contain highly redundant information, and that this redundancy is mainly distributed across frames. However, eliminating this redundancy must take into account the structural characteristics of video, so merging only within local temporal windows or within a single frame cannot eliminate the bulk of the redundancy, indicating the necessity of global-level merging.

Existing video token compression research has evolved along two routes: the first is training-free heuristic methods, which perform unconditional reduction via hard dropping~\citep{chen2024fastv, xing2025pyramiddrop} or equal-weight averaging~\citep{bolya2023tome}. Such methods incur no extra training cost, but their combination weights are fixed, lack the ability to adapt to different scene content, and tend to impair the model's understanding of complex dynamic features. The second is learnable compression methods, which introduce additional parametric modules to re-encode the visual input~\citep{li2023blip2, alayrac2022flamingo}. Such methods have strong fitting capacity, but often face heavy alignment training, and the reconstructed representation deviates from the original distribution the multimodal model was pre-trained on. In addition, a few query-conditioned compression attempts~\citep{huang2025prunevid, li2025dytok} can improve accuracy under a single question, but their compressed representation is tied to a specific query, preventing the KV cache from being reused across multi-turn dialogue. In summary, how to endow the compression operator with content adaptivity while maintaining low training overhead and general-purpose reusability remains a challenge that this field urgently needs to solve.

To alleviate the above dilemma, we propose our solution \textbf{CRAFT (Compression via Recursive Adaptive Fusion of Video Tokens)}, which balances adaptive aggregation and distribution preservation while keeping a low level of computational complexity. Specifically, CRAFT adopts a multi-round iterative ``select-then-fuse'' mechanism, decoupling the compression process within each round into two parts: \emph{which tokens to merge}, which is dynamically decided by training-free global feature similarity. It introduces a process that is parameter-free and query-agnostic, ensuring the cross-round robustness of the compressed representation; and \emph{how to combine}, which is adaptively determined by an end-to-end learnable gated merging mechanism, including a self-attention global weight for all selected tokens and a per-channel local gate for each token individually.

Our contributions are as follows:
(1) To the best of our knowledge, this is the first time an end-to-end iterative trainable ``select-then-fuse'' approach has been introduced for general-purpose visual token compression. The combination of parameter-free selection and parametric aggregation ensures the method's adaptability, robustness, and low computational complexity.
(2) We design a dual-path trainable aggregation mechanism. For a specific selected cluster, this mechanism simultaneously incorporates adaptive capabilities across both token and channel dimensions through a combination of coarse-grained global attention weighting and fine-grained local gating control.
(3) Systematic experiments on six representative video-understanding benchmarks show that CRAFT achieves state-of-the-art performance among the compared methods across compression ratios from $2\times$ to $32\times$, while delivering the lowest prefill latency and the largest FLOPs reduction among all compared methods.

\section{Related Work}

\subsection{Training-Free Token Reduction}

Training-free compression methods shrink the visual token sequence with zero added parameters and zero training cost, offering strong plug-and-play deployability. By how they treat the information of removed tokens, they split into two routes: selective pruning and merging-based aggregation.

Selective pruning focuses on hard-dropping redundant tokens. In terms of pruning schedule, FastV~\citep{chen2024fastv} performs one-shot pruning using attention scores at a shallow layer of the model; PyramidDrop~\citep{xing2025pyramiddrop} instead improves this into staged, progressive dropping along network depth. In terms of the ranking criterion, DivPrune~\citep{alvar2025divprune} introduces max--min diversity to improve the spatial coverage of retained tokens; VidCom$^2$~\citep{liu2025vidcom2} further extends the uniqueness measure to the temporal dimension, achieving dual screening by both frame-level and intra-frame uniqueness.

Merging-based aggregation mitigates information loss by merging similar tokens. The foundational ToMe~\citep{bolya2023tome} uses bipartite matching to merge the most similar token pairs in linear time. For video tasks, subsequent work organizes merging within different spatio-temporal scopes: DyCoke~\citep{tao2025dycoke} and FastVID~\citep{shen2025fastvid} focus on dynamic merging along the temporal dimension; DynTok~\citep{zhang2025dyntok} and LLaVA-Scissor~\citep{sun2025llavascissor} perform adaptive regional aggregation based on information density and semantic connected components, respectively; STTM~\citep{hyun2025sttm}, HoliTom~\citep{shao2025holitom}, and VisionZip~\citep{yang2025visionzip} extend the merging scope to the full spatio-temporal domain, coordinating merging inside and outside the language model with multi-granularity or redundancy-aware mechanisms.

Although the above work keeps refining the scope of selection, the information of dropped tokens is always lost outright, and fusion weights are typically fixed as static rules at design time, limiting content-adaptive trade-offs, and risking irreversible semantic loss at high compression ratios. In contrast, while retaining the low-overhead advantage of training-free similarity-based selection, CRAFT builds a multi-round iterative ``select-then-fuse'' mechanism that shifts the core toward learnable, channel-wise gated fusion weights.

\subsection{Learnable Compression and Conditioned Methods}

Learnable compression methods raise the ceiling of visual representation by introducing trainable parameters~\citep{rao2021dynamicvit, liang2022evit}. Early connectors such as Q-Former~\citep{li2023blip2}, the Perceiver Resampler~\citep{alayrac2022flamingo}, and TokenPacker~\citep{li2025tokenpacker} project arbitrary-length visual sequences into a fixed number of latent tokens. Later designs such as PVC~\citep{yang2025pvc}, LaCo~\citep{liu2025laco}, and InternVL-X~\citep{lu2025internvlx} embed the compressor deep inside the vision encoder or the shallow language-model layers. These designs deviate from the original pre-training input distribution, requiring costly re-alignment training. Beyond such heavyweight re-encoders, recent learnable compression is predominantly query-conditioned, and lightweight query-agnostic learnable compressors have received far less attention: even when token selection is learned with lightweight differentiable modules, as in VisionSelector~\citep{zhu2025visionselector}, fusion still follows a static, equal-weight rule. Query-conditioned methods instead treat the question as a prior, filtering the most task-relevant visual tokens with text-guided attention~\citep{zhang2025sparsevlm, huang2025prunevid, li2025dytok, shen2025longvu} or retrieving question-relevant clips via reinforcement learning~\citep{wu2026marc}. Such compression works well within a single question-answering instance, but the representation is tightly bound to that query, so the compressed KV cache cannot be reused across questions or multi-turn dialogue.

Unlike the methods above, CRAFT restricts the learning target to linear combinations of the original tokens, using iterative selection and fusion to keep the compressed representation within the original pre-training space, avoiding costly re-alignment overhead. Furthermore, CRAFT adopts a query-agnostic compression scheme, allowing the compressed visual representation to be computed once and reused for general purposes, making it effective for multi-turn inference.

\begin{figure*}[t]
    \centering
    \includegraphics[width=1\textwidth]{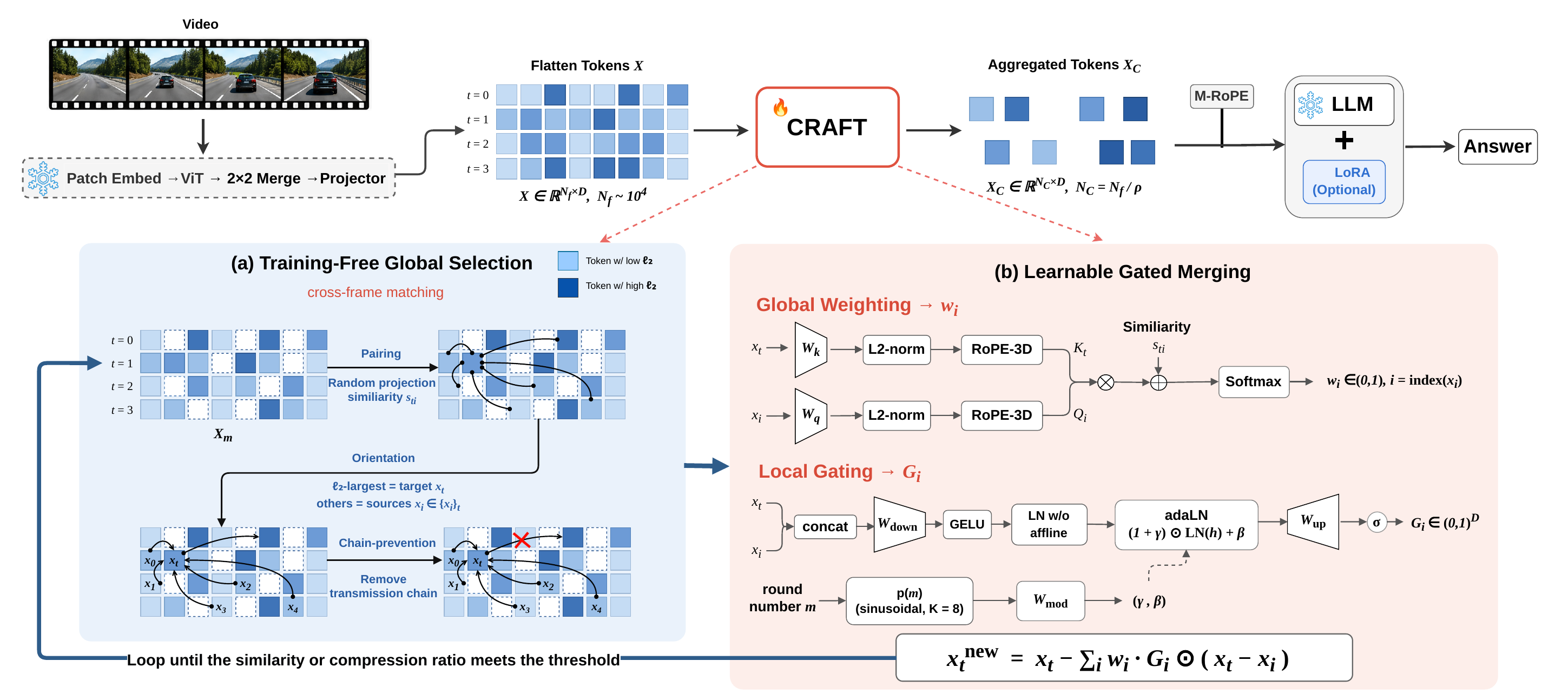}
    \caption{CRAFT performs Training-Free Global Selection and Learnable Gated Merging iteratively. Note that in Training-Free Global Selection, previously merged tokens are shown as white rectangles. For clarity, only one representative token $x_t$ and its associated token set $\{x_i\}_t$ are illustrated.}
    \label{fig:arch}
\end{figure*}

\section{Method}

\subsection{Problem Definition and Overview}
\label{sec:overview}

As shown in Fig.~\ref{fig:arch}, given a video input, the vision encoder and projector project it into the language model's embedding space, producing a visual feature sequence $X \in \mathbb{R}^{N_f \times D}$. In video scenarios, $N_f$ is typically on the order of $10^4$. CRAFT acts on the feature sequence $X$, iteratively identifying pairs of similar tokens and adaptively merging them, ultimately producing a compressed sequence $X_C \in \mathbb{R}^{N_c \times D}$. The iterative process supports two budget modes: 1) an \emph{adaptive similarity-threshold} mode, which stops once the maximum pairwise similarity among retained tokens falls below a threshold $s_{\theta}$, so the realized ratio $\rho = N_f/N_c$ adapts to the video content, with a budget floor on the retained token count, $N_{\min} = \max(1, \lfloor N_f/128 \rfloor)$ in our experiments, preventing over-merging; 2) a \emph{fixed-ratio} mode, which continues until $\rho$ reaches a target ${\rho}_{\theta}$ (Algorithm~\ref{alg:lgm} details the former; the latter is in Appendix~\ref{app:algorithm}).

Notably, CRAFT performs iterative token merging without conditioning on textual queries.
As a result, a video needs to be compressed only once, and the resulting compact visual sequence $X_C$ can be directly cached as visual key-value representations and reused across multiple dialogue turns and diverse user queries:
\begin{equation}\label{eq:objective}
p(\mathbf{y} \mid X_C, \mathbf{q}) = \prod_{\ell=1}^{L} p(y_\ell \mid y_{<\ell}, X_C, \mathbf{q})
\end{equation}

\subsection{Training-Free Global Selection}
\label{sec:selection}

Since redundancy in videos often spans across frames,
intra-frame or local sliding-window merging strategies cannot fundamentally eliminate the majority of redundant information.
Therefore, token selection must be performed at a global scale.
To maintain extremely low computational overhead on high-dimensional sequences at the $N_f \sim 10^4$ level, we exploit the inner-product/cosine-similarity-preserving property of random projections~\citep{johnson1984lipschitz, achlioptas2003database}, computing similarity in a fixed low-dimensional random projection space:
\begin{equation}\label{eq:sim}
E_{\text{sim}} = \overline{X W_{\text{rand}}}, \qquad W_{\text{rand}} \in \mathbb{R}^{D \times d_{\text{sim}}}.
\end{equation}
Here $\overline{\cdot}$ denotes $\ell_2$ normalization, and $d_{\text{sim}}$ is typically set to 128. The projection matrix $W_{\text{rand}}$ is generated once with a fixed random seed and column-normalized. The similarity score between any two tokens $i,j$ is given directly by the inner product of their projected unit vectors:
\begin{equation}\label{eq:cossim}
s_{ij} = E_{\text{sim},i} E_{\text{sim},j}^\top \approx \cos(x_i, x_j)
\end{equation}
Similarity computation is carried out via blockwise matrix multiplication with block size $2048$ to prevent extreme GPU memory peaks.

Each iteration consists of four stages: pairing, budget truncation, orientation, and chain prevention (Algorithm~\ref{alg:lgm}).
In the pairing stage, each active token nominates its most similar token, $p(i) = \arg\max_{j \neq i} s_{ij}$, yielding one candidate edge $(i, p(i))$ per token with similarity $s_i^{\star} = s_{i,p(i)}$.
The budget-truncation stage then merges only the $k = \min(|\text{active}| - N_{\min}, \lfloor |\text{active}|/2 \rfloor)$ highest-similarity edges in the current round (edges with $s_i^{\star} < s_{\theta}$ are further excluded in the adaptive threshold mode), keeping the retained count above the budget floor $N_{\min}$.
Next, during the orientation stage, the token with the larger $\ell_2$-norm in each selected edge is retained as the representative token, denoted by $x_t$, based on the assumption that a larger norm indicates more salient semantic information.
As a result, multiple tokens may be associated with the same representative token $x_t$, forming a group $\{x_i\}_t$.
However, some representative tokens may themselves be selected as the source of another edge whose partner has an even larger $\ell_2$-norm.
To avoid such chain-style aggregation, these cases are resolved in the chain-prevention stage: rather than being discarded, a conflicting representative token simply drops its outgoing edge, is excluded from merging in the current iteration, and remains active, re-entering the pairing stage in the next iteration. The surviving representative token preserves its original spatio-temporal coordinates $(t,h,w)$, thereby maintaining positional information throughout the merging process.

\begin{algorithm}[!t]
\caption{CRAFT Select-then-Fuse (adaptive mode)}
\label{alg:lgm}
\small
\begin{algorithmic}[1]
\REQUIRE Features $X$, coords $\{(t,h,w)\}$, threshold $s_{\theta}$, budget floor $N_{\min}$
\ENSURE Compressed sequence $X_C$, updated coords
\STATE $\text{active} \leftarrow \{1,\dots,N_f\}$; $m \leftarrow 0$; $E_{\text{sim}} \leftarrow \overline{X W_{\text{rand}}}$ \COMMENT{\textcolor{gray}{$\triangleright$ Eq. 2}}
\WHILE{$|\text{active}| > N_{\min}$}
    \STATE $m \leftarrow m + 1$
    \STATE $p(i) \leftarrow \arg\max_{j \neq i} s_{ij}$;\; $s_i^{\star} \leftarrow s_{i,p(i)}$,\; $\forall i \in \text{active}$ \COMMENT{\textcolor{gray}{$\triangleright$ Pairing (Eq. 3)}}
    \STATE $C \leftarrow \{(i, p(i)) : s_i^{\star} \geq s_{\theta}\}$;\; \textbf{if} $C = \emptyset$ \textbf{then break} \COMMENT{\textcolor{gray}{$\triangleright$ Adaptive stop}}
    \STATE $k \leftarrow \min\!\big(|\text{active}| - N_{\min},\, \lfloor |\text{active}|/2 \rfloor,\, |C|\big)$
    \STATE $P \leftarrow$ the $k$ edges of $C$ with highest $s_i^{\star}$ \COMMENT{\textcolor{gray}{$\triangleright$ Budget truncation}}
    \STATE Orient edges ($\ell_2$-larger endpoint $\to$ target), drop chain-conflicting edges, group into $(t, \{i\}_t)$ \COMMENT{\textcolor{gray}{$\triangleright$ \S3.2}}
    \FORALL{merge group $(t, \{i\}_t)$}
        \STATE $x_t^{\text{new}} \leftarrow \text{GatedMerge}(x_t, \{x_i\}_t)$ \COMMENT{\textcolor{gray}{$\triangleright$ Eqs. 4--9}}
    \ENDFOR
    \STATE $X[\text{targets}] \leftarrow x_t^{\text{new}}$; $\text{active} \leftarrow \text{active} \setminus \{\text{sources}\}$
    \STATE $E_{\text{sim}}[\text{targets}] \leftarrow \overline{X[\text{targets}] W_{\text{rand}}}$ \COMMENT{\textcolor{gray}{$\triangleright$ Refresh (Eq. 2)}}
\ENDWHILE
\RETURN $X_C \leftarrow X[\text{active}]$ with updated coordinates
\end{algorithmic}
\end{algorithm}

\subsection{Learnable Gated Merging}
\label{sec:merging}

\textit{How are the selected pairs combined?} For each representative $x_t$ and its source set $\{x_i\}$, CRAFT fuses by
\begin{equation}\label{eq:fuse}
x_t^{\text{new}} = x_t - \sum_i w_i\, G_i \odot (x_t - x_i).
\end{equation}
$w_i$ is the weight among the sources, satisfying $\sum_i w_i = 1$, thereby enabling the aggregation of information from a global perspective. $G_i \in (0,1)^D$ denotes the per-channel gating vector for the local perspective, and $\odot$ denotes the Hadamard product. When $G_i \equiv 1$, Eq.~(\ref{eq:fuse}) reduces to $x_t^{\text{new}} =  \Bigl(1-\sum_i w_i\Bigr)x_t + \sum_i w_i x_i = \sum_i w_i x_i$ , i.e.\ it degenerates into a weighted average of the source tokens; if $w_i$ is further taken to be uniform, this becomes the traditional equal-weight average. When $G_i \equiv 0$, $\sum_i w_i G_i \odot (x_t - x_i) \equiv 0$, and Eq.~(\ref{eq:fuse}) degenerates into hard-dropping the sources and keeping only the representative.
This design allows the contribution of each token to the merging process to be adaptively regulated from both global and local perspectives.

\textbf{Global position-aware weighting (source weights $w_i$).} The weighting scorer first models spatio-temporal relations through low-rank projection and 3D positional rotation:
\begin{equation}\label{eq:qk}
\hat{Q} = \mathrm{RoPE}_{3d}(\overline{X W_q}), \quad \hat{K} = \mathrm{RoPE}_{3d}(\overline{X W_k}).
\end{equation}
Here $W_q, W_k \in \mathbb{R}^{D \times r}$, $r=128$, with the $\mathrm{RoPE}_{3d}$ following M-RoPE~\citep{wang2024qwen2vl}.
The positional score is computed by the projected dot product between the query and key vectors on the selected merge pair.
Then the similarity score $s_{ti}$ from selection is added to the positional score.
The weights among sources are adaptively assigned via softmax normalization over the representative token's source set:
\begin{equation}\label{eq:weight}
w_i = \frac{\exp((\hat{Q}_i\hat{K}_t^{\top} + s_{ti})/\tau)}{\sum_{j} \exp((\hat{Q}_j\hat{K}_t^{\top} + s_{tj})/\tau)}.
\end{equation}

\textbf{Local content-conditioned gating (per-channel $G_i$).} The gating network produces $G_i$ of the same size as $x_t$, enabling fine-grained control over the contribution of each channel during token merging. It takes the concatenation of the representative feature and the source feature as input, and, via adaptive layer normalization (adaLN) incorporating the merge round $m$, outputs a multi-channel adaptive gating vector:
\begin{equation}\label{eq:gatehidden}
h_i = \mathrm{GELU}\bigl(W_{\text{down}}\,[\,x_t;\,x_i\,]\bigr) \in \mathbb{R}^{d_h},
\end{equation}
\begin{equation}\label{eq:adaln}
\tilde{h}_i = (1 + \gamma_i) \odot \mathrm{LN}(h_i) + \beta_i, \quad (\gamma_i, \beta_i) = W_{\text{mod}}\, \phi(m) + b_{\text{mod}},
\end{equation}
\begin{equation}\label{eq:gate}
G_i = \sigma\bigl(W_{\text{up}}\, \tilde{h}_i\bigr) \in (0,1)^D.
\end{equation}
Here $d_h$ is typically 512, $\sigma$ denotes the Sigmoid activation function, and $\mathrm{LN}$ denotes layer normalization without affine parameters. $\phi(m)$
denotes the sine-cosine positional encoding of round $m$, which provides the network with awareness of the current merging stage. Notably, the linear projection $W_{\text{up}}$, as well as the modulation projections $W_{\text{mod}}, b_{\text{mod}}$, adopt zero initialization (adaLN-Zero~\citep{peebles2023dit}), so that the network's initial output is $\gamma_i=\beta_i=0$ and $G_i=0.5$.

\textbf{Overall procedure.} The training-free selection of \S\ref{sec:selection} and the learnable fusion above are not two sequential stages but two sub-steps within the same iterative loop, in which each round first selects the pairs to be merged, then performs the weighted fusion, and refreshes the similarities before entering the next round based on the merged tokens, until the compression budget is reached. Algorithm~\ref{alg:lgm} gives the complete pseudocode of this iterative ``select-then-fuse'' loop (adaptive threshold mode; the fixed-ratio variant is in Appendix~\ref{app:algorithm}).

\subsection{Two-Stage Training Strategy}
\label{sec:training}

We freeze the vision encoder, multimodal projector, and large language model, and optimize only the weighting scorer and gating network under the standard cross-entropy objective, so training stays lightweight and the backbone's input distribution remains untouched.
Training follows a two-stage curriculum.
The first stage pre-trains the gate on multi-source video-captioning data, encouraging it to retain globally relevant semantic information during compression.
The second stage continues on video question-answering (QA) data, aligning the compressor with downstream reasoning objectives and steering it toward fine-grained visual details.
The data composition of both stages is detailed in Appendix~\ref{app:data}.

\section{Experiments}

\begin{table*}[t]\centering
\small
\setlength{\tabcolsep}{2pt} %
\newcommand{\rt}[1]{\enskip{\scriptsize(#1)}} %
\newcommand{\cc}[2]{\begin{tabular}[c]{@{}c@{}}#1\\[-1.5pt]{\scriptsize(#2)}\end{tabular}} %

\begin{tabular}{ll c ccccccc}
\toprule
\textbf{Backbone} & \textbf{Method} & \textbf{Comp.} & \textbf{Video-MME} & \textbf{LongVideoBench} & \textbf{MVBench} & \textbf{EgoSchema} & \textbf{MLVU} & \textbf{TempCompass} & \textbf{Avg.} \\

\midrule
\multicolumn{10}{l}{\textbf{\textit{A. Main Comparison}}} \\
\midrule
\multirow[t]{8}{*}{\textbf{Qwen3.5-4B}}
 & \color{basegray}Base & \color{basegray}$1\times$ & \color{basegray}\cc{67.0}{100\%} & \color{basegray}\cc{58.4}{100\%} & \color{basegray}\cc{68.4}{100\%} & \color{basegray}\cc{67.0}{100\%} & \color{basegray}\cc{69.3}{100\%} & \color{basegray}\cc{74.3}{100\%} & \color{basegray}\cc{67.4}{100\%} \\
 & VisionZip         & $8\times$ & \cc{61.5}{91.8\%} & \cc{52.1}{89.2\%} & \cc{59.1}{86.4\%} & \cc{63.4}{94.6\%} & \cc{63.3}{91.3\%} & \cc{67.9}{91.3\%} & \cc{61.2}{90.8\%} \\
 & FastVID           & $8\times$ & \cc{60.2}{89.9\%} & \cc{54.2}{92.8\%} & \cc{59.9}{87.5\%} & \cc{60.0}{89.6\%} & \cc{64.8}{93.4\%} & \cc{\underline{69.4}}{\underline{93.4\%}} & \cc{61.4}{91.1\%} \\
 & DivPrune          & $8\times$ & \cc{\underline{63.7}}{\underline{95.1\%}} & \cc{54.2}{92.8\%} & \cc{61.2}{89.5\%} & \cc{63.0}{94.0\%} & \cc{\underline{66.2}}{\underline{95.4\%}} & \cc{68.2}{91.7\%} & \cc{\underline{62.7}}{\underline{93.1\%}} \\
 & VidCom\textsuperscript{2} & $8\times$ & \cc{62.5}{93.3\%} & \cc{\underline{56.0}}{\underline{95.8\%}} & \cc{59.9}{87.5\%} & \cc{\underline{63.8}}{\underline{95.2\%}} & \cc{64.3}{92.8\%} & \cc{65.5}{88.2\%} & \cc{62.0}{92.0\%} \\
 & LLaVA-Scissor     & $8\times$ & \cc{62.3}{93.0\%} & \cc{53.2}{91.0\%} & \cc{\underline{62.8}}{\underline{91.7\%}} & \cc{63.4}{94.6\%} & \cc{64.9}{93.6\%} & \cc{69.3}{93.3\%} & \cc{62.6}{92.9\%} \\
 \rowcolor{gamebg} & \textbf{CRAFT (Ours)} & $\boldsymbol{8\times}$ & \cc{\textbf{64.6}}{\textbf{96.4\%}} & \cc{\textbf{58.2}}{\textbf{99.6\%}} & \cc{\textbf{63.0}}{\textbf{92.1\%}} & \cc{\textbf{65.8}}{\textbf{98.2\%}} & \cc{\textbf{70.0}}{\textbf{101.0\%}} & \cc{\textbf{70.0}}{\textbf{94.2\%}} & \cc{\textbf{65.3}}{\textbf{96.8\%}} \\
 \rowcolor{gamebg} & \color{basegray}CRAFT+ (Ours)$^{\ddagger}$ & \color{basegray}$10\times$ & \color{basegray}\cc{64.0}{95.5\%} & \color{basegray}\cc{59.5}{101.8\%} & \color{basegray}\cc{66.2}{96.8\%} & \color{basegray}\cc{70.8}{105.7\%} & \color{basegray}\cc{69.0}{99.5\%} & \color{basegray}\cc{71.8}{96.6\%} & \color{basegray}\cc{66.9}{99.2\%} \\

\midrule
\multicolumn{10}{l}{\textbf{\textit{B. Cross-Backbone Transfer}}} \\
\midrule
\multirow[t]{5}{*}{\textbf{Qwen2.5-VL-7B}}
 & \color{basegray}Base & \color{basegray}$1\times$ & \color{basegray}\cc{64.1}{100\%} & \color{basegray}\cc{60.3}{100\%} & \color{basegray}\cc{67.8}{100\%} & \color{basegray}\cc{64.6}{100\%} & \color{basegray}\cc{67.4}{100\%} & \color{basegray}\cc{72.6}{100\%} & \color{basegray}\cc{66.1}{100\%} \\
 & TimeChat-Online & $7.3\times$ & \cc{\underline{59.9}}{\underline{93.4\%}} & \cc{\underline{54.4}}{\underline{90.2\%}} & \cc{\textbf{63.9}}{\textbf{94.2\%}} & \cc{50.0}{77.4\%} & \cc{\underline{62.3}}{\underline{92.4\%}} & \cc{\underline{68.3}}{\underline{94.1\%}} & \cc{59.8}{90.5\%} \\
 & VisionSelector & $8\times$ & \cc{\textbf{60.0}}{\textbf{93.6\%}} & \cc{54.2}{89.9\%} & \cc{62.9}{92.8\%} & \cc{\textbf{65.8}}{\textbf{101.9\%}} & \cc{59.6}{88.4\%} & \cc{65.4}{90.1\%} & \cc{\underline{61.3}}{\underline{92.7\%}} \\
  \rowcolor{gamebg} & \textbf{CRAFT (Ours)} & $\boldsymbol{8\times}$ & \cc{\underline{59.9}}{\underline{93.4\%}} & \cc{\textbf{57.0}}{\textbf{94.5\%}} & \cc{\underline{63.8}}{\underline{94.1\%}} & \cc{\underline{61.8}}{\underline{95.7\%}} & \cc{\textbf{66.2}}{\textbf{98.2\%}} & \cc{\textbf{69.3}}{\textbf{95.5\%}} & \cc{\textbf{63.0}}{\textbf{95.3\%}} \\
\midrule
\multirow[t]{3}{*}{\textbf{LLaVA-OV-7B}}
 & \color{basegray}Base & \color{basegray}$1\times$ & \color{basegray}\cc{59.0}{100\%} & \color{basegray}\cc{57.0}{100\%} & \color{basegray}\cc{56.9}{100\%} & \color{basegray}\cc{64.4}{100\%} & \color{basegray}\cc{67.2}{100\%} & \color{basegray}\cc{63.7}{100\%} & \color{basegray}\cc{61.4}{100\%} \\
 \rowcolor{gamebg} & \textbf{CRAFT (Ours)} & $\boldsymbol{8\times}$ & \cc{\textbf{55.5}}{\textbf{94.1\%}} & \cc{\textbf{52.4}}{\textbf{91.9\%}} & \cc{\textbf{56.0}}{\textbf{98.4\%}} & \cc{\textbf{63.4}}{\textbf{98.4\%}} & \cc{\textbf{63.9}}{\textbf{95.1\%}} & \cc{\textbf{62.9}}{\textbf{98.7\%}} & \cc{\textbf{59.0}}{\textbf{96.1\%}} \\
\bottomrule
\end{tabular}
\caption{Main results on six video benchmarks. Below each accuracy is its retention relative to the corresponding uncompressed backbone. The ``Comp.'' column is the average realized token compression ratio across benchmarks. \textbf{Block~A} compares CRAFT under adaptive mode with training-free token-reduction methods on Qwen3.5-4B; \textbf{Block~B} transfers CRAFT to Qwen2.5-VL-7B and LLaVA-OV-7B. \textbf{Bold}/\underline{underline} mark the best/second-best compressed result per column. CRAFT is our original implementation. $^{\ddagger}$CRAFT$+$ additionally adapts the LLM with LoRA and is reported in gray for reference only. \textbf{Full accuracy--compression trade-off from $2\times$ to $32\times$ on Qwen3.5-4B is given in Fig.~\ref{fig:tradeoff}.} }
\label{tab:exp_main}
\end{table*}

\subsection{Experimental Setup}
\label{sec:exp_setup}

\noindent\textbf{Benchmarks.} We evaluate on six video-understanding benchmarks that span short clips to long-form temporal reasoning, including Video-MME~\citep{fu2025videomme} (w/o subtitles), LongVideoBench~\citep{wu2024longvideobench}, MVBench~\citep{li2024mvbench}, EgoSchema~\citep{mangalam2023egoschema}, MLVU~\citep{zhou2024mlvu}, and TempCompass~\citep{liu2024tempcompass}. Unless otherwise noted, all numbers are top-1 accuracy under greedy decoding, sampling frames at $\mathrm{fps}{=}2$ up to a maximum of $64$ frames, on the full test/validation splits; benchmark-specific evaluation details for MVBench and TempCompass are given in Appendix~\ref{app:benchmarks}.

\noindent\textbf{Metrics.} Besides raw accuracy we report the retention $\mathrm{acc}/\mathrm{acc}_{\mathrm{base}}$ relative to the corresponding uncompressed backbone, shown as the small parenthetical below each accuracy in Table~\ref{tab:exp_main}, and the realized token-compression ratio $\rho{=}N_f/N_c$ (computed per video and averaged over each dataset), shown as the ``Comp.'' column. Table~\ref{tab:exp_main}, the efficiency analysis of \S\ref{sec:efficiency}, and Fig.~\ref{fig:gate-vs-scissor} evaluate CRAFT under the adaptive similarity-threshold mode, whereas the remaining experiments (Fig.~\ref{fig:teaser} and Table~\ref{tab:exp_ablation}) use the fixed-ratio mode to keep the token budget identical across configurations.

\noindent\textbf{Backbones.} Qwen3.5-4B~\citep{qwenteam2026qwen35omnitechnicalreport} is our primary backbone. To demonstrate that CRAFT is architecture-agnostic, we transfer it to Qwen2.5-VL-7B~\citep{bai2025qwen25vl} and LLaVA-OV-7B~\citep{li2024llavaonevisioneasyvisualtask}, the two most widely adopted architectures in token-compression studies, which differ in vision encoder and token layout.

\noindent\textbf{Baselines.} On Qwen3.5-4B we compare against five representative training-free token-reduction methods covering both the pruning and the merging routes, DivPrune~\citep{alvar2025divprune}, VidCom$^2$~\citep{liu2025vidcom2}, LLaVA-Scissor~\citep{sun2025llavascissor}, VisionZip~\citep{yang2025visionzip}, and FastVID~\citep{shen2025fastvid}, all transferred plug-and-play onto our backbone so that every baseline is compared on an identical footing. On Qwen2.5-VL-7B we additionally compare against two learnable compressors, TimeChat-Online~\citep{yao2025timechatonline} and VisionSelector~\citep{zhu2025visionselector}. Because VisionSelector relies on a trainable compressor module and TimeChat-Online fine-tunes its backbone on dedicated data, porting them to another backbone would introduce confounds, so both are kept on their native Qwen2.5-VL-7B. All methods are run under the same frame budget for fairness.

\noindent\textbf{Implementation.} We build CRAFT, and its LoRA-adapted variant CRAFT$+$, on Qwen3.5-4B and transfer the same compressor design unchanged to Qwen2.5-VL-7B and LLaVA-OV-7B. The training curriculum and hyperparameters are detailed in Appendix~\ref{app:hparams}.

\subsection{Main Results}
\label{sec:main_results}

\noindent\textbf{Comparison with training-free methods.} Table~\ref{tab:exp_main} (block~A) reports the head-to-head comparison of CRAFT against a range of state-of-the-art training-free methods on Qwen3.5-4B, and the results confirm the advantage of a \emph{learnable} gated fusion over the fixed, heuristic rules that every training-free baseline relies on. At ${\sim}8\times$ compression CRAFT retains $96.8\%$ of the backbone's average accuracy ($65.3$ vs.\ $67.4$), outperforming every training-free baseline by a clear margin. The strongest competitor, DivPrune, reaches only $93.1\%$, while VisionZip and FastVID fall to ${\sim}91\%$.
Notably, CRAFT even surpasses the uncompressed backbone on MLVU, reaching $101.0\%$ of the original accuracy.
Adaptive thresholding grants CRAFT no extra budget: the dataset-averaged realized ratio is $8.0\times$, matching the baselines, and tokens are merely reallocated from redundant clips to information-dense ones---itself an advantage of content-aware compression. Even at a strictly fixed $8\times$ budget, CRAFT averages $64.8$ (Table~\ref{tab:exp_ablation}, last row), still clearly ahead of the best training-free baseline ($62.7$).
The superiority of CRAFT extends beyond any single benchmark.
Under a fixed $8\times$ compression budget, it also achieves the best or equal-level accuracy across all six benchmarks in Fig.~\ref{fig:teaser}.

\begin{figure}[t]
    \centering
    \includegraphics[width=\columnwidth]{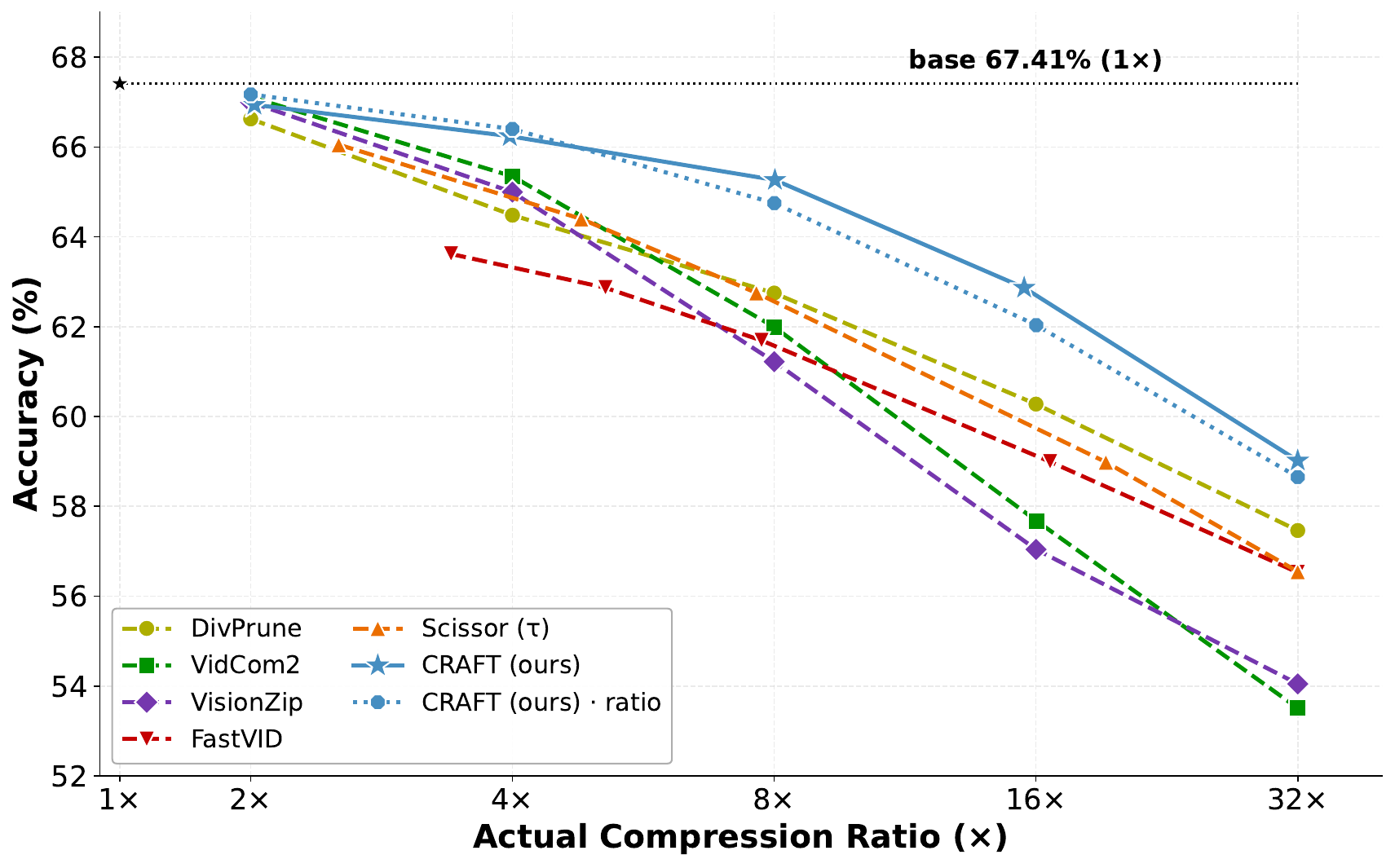}
    \caption{Accuracy--compression trade-off on Qwen3.5-4B: benchmark-averaged top-1 accuracy vs.\ realized compression ratio ($2\times$ to $32\times$).}
    \label{fig:tradeoff}
\end{figure}

\noindent\textbf{Accuracy--compression trade-off.}
While a single operating point may favor a particular method, the rate of accuracy degradation under tighter token budgets is more telling.
Fig.~\ref{fig:tradeoff} sweeps the compression ratio from $2\times$ to $32\times$ on Qwen3.5-4B.
Up to $4\times$ every method stays close to the uncompressed base of $67.41\%$, but beyond ${\sim}8\times$ the training-free baselines decline sharply toward $53.5$--$58\%$ at $32\times$, whereas CRAFT degrades gracefully, widens its margin as the ratio grows, and still reaches nearly $59\%$.
The steadily growing margin shows that learnable gated fusion, rather than any static merging rule, is what preserves critical information when the token budget becomes severely constrained.

\subsection{Cross-Backbone Transfer}
\label{sec:transfer}

Without any change to the compressor design, CRAFT transfers to two further backbones (Table~\ref{tab:exp_main}, block~B; porting details in Appendix~\ref{app:transfer}).
On Qwen2.5-VL-7B, it retains $95.3\%$ of the full-model performance ($63.0$) at an $8\times$ compression ratio, outperforming compressors trained on the same backbone, including VisionSelector ($92.7\%$) and TimeChat-Online ($90.5\%$), while requiring the training of only a lightweight ${\sim}5$M-parameter module.

On LLaVA-OV-7B, CRAFT retains $96.1\%$ of the full-model performance.
That the same design generalizes across three architectures with distinct vision encoders and token layouts indicates that its effectiveness arises from the decoupling of training-free selection and learnable, position-aware fusion, rather than architecture-specific tuning.

\subsection{Efficiency Analysis}
\label{sec:efficiency}

Token compression aims to cut the cost of the prefill stage, where visual tokens dominate the input at $N_f \sim 10^4$. CRAFT itself adds little overhead. Pair selection is linear in the surviving tokens per round, and similarity is computed in a fixed $d_{\text{sim}}{=}128$ random-projection space with blockwise products, so the full $N_f \times N_f$ matrix is never materialized. Its two learnable modules hold only ${\sim}5$M parameters, negligible beside a $4$B-parameter LLM, so the select-then-fuse step costs far less than it saves downstream.

\begin{table}[t]\centering
\small
\setlength{\tabcolsep}{3pt}
\renewcommand{\arraystretch}{1.15}
\begin{tabular}{l c c c c}
\toprule
Method & Acc.(\%) & Prefill (ms)$\downarrow$ & KV (MB)$\downarrow$ & FLOPs$\uparrow$ \\
\midrule
\color{basegray}Base & \color{basegray}68.4 & \color{basegray}249.2 & \color{basegray}162.5 & \color{basegray}$1\times$ \\
DivPrune             & 61.2 & 72.3 ($\uparrow$11.6\%) & 26.0 & $6.75\times$ \\
VidCom$^2$           & 59.9 & 72.6 ($\uparrow$12.0\%) & 26.0 & $6.75\times$ \\
VisionZip            & 59.1 & 72.5 ($\uparrow$11.9\%) & 26.0 & $6.75\times$ \\
FastVID              & 59.9 & 72.8 ($\uparrow$12.3\%) & 26.0 & $6.75\times$ \\
LLaVA-Scissor        & 62.8 & 70.5 ($\uparrow$8.8\%)  & 26.2 & $6.71\times$ \\
\rowcolor{gamebg} \textbf{CRAFT} & \textbf{63.2} & \textbf{64.8} & \textbf{19.0} & $\mathbf{9.30\times}$ \\
\bottomrule
\end{tabular}
\caption{Prefill-stage efficiency and analytical FLOPs reduction on MVBench, at the operating point where CRAFT's accuracy first matches the strongest merging-based baseline, LLaVA-Scissor.}
\label{tab:efficiency}
\end{table}

Table~\ref{tab:efficiency} reports this at the point where CRAFT first matches LLaVA-Scissor, the strongest merging-based baseline on this benchmark. CRAFT reaches $63.2\%$, surpassing every baseline while leading on all three efficiency axes.

\noindent\textbf{Time.} CRAFT reaches the lowest prefill latency of any method compared, $8.8$--$12.3\%$ faster than every training-free baseline, whose own mutual spread is only $3.4\%$.

\noindent\textbf{Space.} The per-sample KV cache shrinks to $19.0$MB, $27$--$28\%$ smaller than every training-free baseline's $26.0$--$26.2$MB---the budget every additional dialogue turn reuses without recomputation.

\noindent\textbf{FLOPs.} The analytical reduction of FLOPs, computed purely from the surviving token count and architecture, reaches $9.30\times$ for CRAFT versus $6.71$--$6.75\times$ for training-free baselines---the largest computational saving of any method compared.

\begin{table}[t]\centering
\small
\setlength{\tabcolsep}{5pt}
\renewcommand{\arraystretch}{1.15}
\begin{tabular}{l l c c c}
\toprule
Weight & Reduce & Stage & Avg. & $\Delta$ \\
\midrule
mean & mean & 1 & 61.81 & -- \\
mean & target & 1 & 62.81 & $\mathbf{+1.00}$ \\
mean & gate & 1 & 63.62 & $\mathbf{+1.81}$ \\
pos-aware & gate & 1 & 63.92 & $\mathbf{+2.11}$ \\
\rowcolor{gamebg} \textbf{pos-aware} & \textbf{gate} & \textbf{2} & \textbf{64.75} & $\mathbf{+2.94}$ \\
\bottomrule
\end{tabular}
\caption{Progressive ablation of \textbf{CRAFT} on Qwen3.5-4B at a fixed $8\times$ ratio, $6$-benchmark average. \textbf{Weight} is how the merged neighbors are weighted across tokens (mean = uniform; pos-aware = the global position-aware weighting of \S\ref{sec:merging}); \textbf{Reduce} is how the weighted neighbors are fused into the surviving token; \textbf{Stage}: 1 = caption pretraining; 2 = + stage-2 QA.}
\label{tab:exp_ablation}
\end{table}

\subsection{Ablation Experiment}
\label{sec:ablation}

Table~\ref{tab:exp_ablation} enables one design choice at a time on Qwen3.5-4B, and the six-benchmark average rises monotonically at every step. Notably, the two fixed reduction rules are exactly the two degenerate extremes of the gate in Eq.~(\ref{eq:fuse}). The \emph{mean} rule corresponds to $G_i \equiv 1$, i.e., equal-weight averaging of the group; the \emph{target} rule corresponds to $G_i \equiv 0$, i.e., pruning. Switching from the former extreme to the latter lifts the average by $1.00$, validating our assumption that tokens with larger $\ell_2$ norm carry more salient semantics. Replacing the two fixed endpoints with the learned gate adds a further $0.81$. The global position-aware weighting of \S\ref{sec:merging} contributes another $0.30$ by reweighting neighbor tokens with content and relative $t,h,w$ position. The stage-2 QA curriculum adds a final $0.83$, yielding the full compressor at $64.75$. The two learnable components act at different granularities: the gate locally per channel, and the weighting scorer globally across tokens. Together they recover most of the accuracy that fixed-rule merging discards.

\subsection{Evaluation of Scaffold and Learnable Merging}
\label{sec:exp-scaffold}

\begin{figure}[t]
    \centering
    \includegraphics[width=\columnwidth]
    {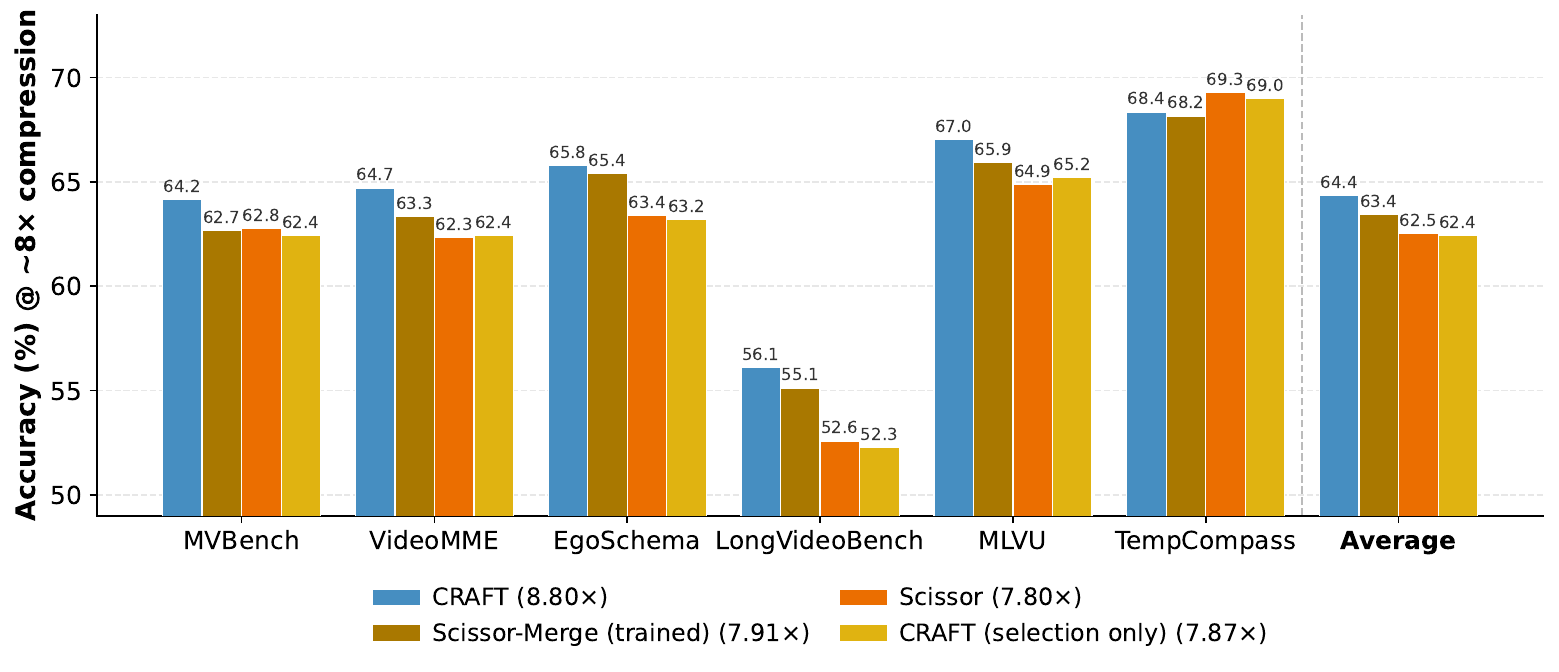}
    \caption{\textbf{Selection scaffold vs.\ learnable merging, at ${\sim}8\times$.} Scissor-Merge couples LLaVA-Scissor's scaffold with our learnable merging module. Scissor is the original LLaVA-Scissor, and CRAFT (selection only) is our scaffold with the learnable merging replaced by equal-weight averaging. All methods run in their threshold-driven adaptive mode, and all trained variants are stage-1-only checkpoints.}
    \label{fig:gate-vs-scissor}
\end{figure}

The compressor of CRAFT involves two separable design choices. The training-free selection scaffold of \S\ref{sec:selection} decides by global feature similarity which tokens are grouped in each round, while the learnable gated merging of \S\ref{sec:merging} decides how each group is fused. To attribute accuracy to each part and to test whether the merging module depends on the scaffold it is paired with, we compare four configurations at a shared ${\sim}8\times$ budget on six benchmarks in Fig.~\ref{fig:gate-vs-scissor}. Besides the full CRAFT and a selection-only CRAFT whose learnable merging is replaced by equal-weight averaging, we evaluate LLaVA-Scissor and a Scissor-Merge variant that feeds our merging module with LLaVA-Scissor's selection module. Scissor-Merge is trained under exactly the same data and hyperparameters as CRAFT, so the two differ only in which scaffold feeds the shared merging module.

The results support three observations. The scaffold alone is already competitive, as the selection-only variant reaches an average of $62.4$, on par with original LLaVA-Scissor ($62.5$). The learnable merging likewise generalizes to a different scaffold, since grafting it onto LLaVA-Scissor's selection lifts the average from $62.5$ to $63.4$, showing that learnable, content-adaptive fusion also improves an established scaffold with a fixed fusion rule. Finally, the two act as complements rather than substitutes: the same merging module that adds $0.9$ on LLaVA-Scissor's scaffold adds $2.0$ on ours ($62.4 \to 64.4$), and pairing it with our own scaffold reaches the best of the four configurations at $64.4$ on average.

\section{Conclusion}

In this paper, we present CRAFT, a query-agnostic visual token compressor for video vision-language models that recursively merges tokens through a decoupled select-then-fuse mechanism, where training-free global similarity decides which tokens to merge and a learnable position-aware weighting with a content-adaptive per-channel gate decides how to fuse them. Extensive experiments show that CRAFT consistently outperforms state-of-the-art compressors and transfers unchanged across three backbones, retaining about $97\%$ of the backbone's average accuracy at roughly $8\times$ compression. CRAFT offers a practical path toward efficient long-video understanding beyond hand-written rules and heavy re-encoding.

\bibliography{aaai2027}

\appendix

\section{Method Details}

\subsection{Algorithm}
\label{app:algorithm}
Algorithm~\ref{alg:lgm} presents the adaptive similarity-threshold mode of CRAFT. Algorithm~\ref{alg:lgm2} below gives the counterpart for the fixed-compression-ratio mode: merging continues until the retained token count reaches the fixed target $N_{\text{tgt}} = \max(1, \lfloor N_f/\rho_{\theta} \rfloor)$, i.e., until the compression ratio $\rho = N_f/N_c$ reaches ${\rho}_{\theta}$. The per-round pairing, orientation, chain prevention, and gated fusion are identical to the threshold mode; the only differences are that no similarity threshold is applied when truncating candidate edges, and the loop terminates solely on the token budget $N_{\text{tgt}}$ rather than on the similarity level, so $N_{\text{tgt}}$ itself plays the role of the budget floor $N_{\min}$ of the threshold mode.

\begin{algorithm}[!t]
\caption{CRAFT Select-then-Fuse (fixed-ratio mode)}
\label{alg:lgm2}
\small
\begin{algorithmic}[1]
\REQUIRE Features $X$, coords $\{(t,h,w)\}$, target ratio ${\rho}_{\theta}$
\ENSURE Compressed sequence $X_C$, updated coords
\STATE $\text{active} \leftarrow \{1,\dots,N_f\}$; $m \leftarrow 0$; $E_{\text{sim}} \leftarrow \overline{X W_{\text{rand}}}$ \COMMENT{\textcolor{gray}{$\triangleright$ Eq. 2}}
\STATE $N_{\text{tgt}} \leftarrow \max(1, \lfloor N_f/\rho_{\theta} \rfloor)$ \COMMENT{\textcolor{gray}{$\triangleright$ Fixed token budget}}
\WHILE{$|\text{active}| > N_{\text{tgt}}$}
    \STATE $m \leftarrow m + 1$
    \STATE $p(i) \leftarrow \arg\max_{j \neq i} s_{ij}$;\; $s_i^{\star} \leftarrow s_{i,p(i)}$,\; $\forall i \in \text{active}$ \COMMENT{\textcolor{gray}{$\triangleright$ Pairing (Eq. 3)}}
    \STATE $k \leftarrow \min\!\big(|\text{active}| - N_{\text{tgt}},\, \lfloor |\text{active}|/2 \rfloor\big)$
    \STATE $P \leftarrow$ the $k$ edges $(i, p(i))$ with highest $s_i^{\star}$ \COMMENT{\textcolor{gray}{$\triangleright$ Budget truncation}}
    \STATE Orient each edge in $P$ ($\ell_2$-larger endpoint $\to$ target), drop chain-conflicting edges, and group into $(t, \{i\}_t)$ \COMMENT{\textcolor{gray}{$\triangleright$ \S3.2}}
    \FORALL{merge group $(t, \{i\}_t)$}
        \STATE $x_t^{\text{new}} \leftarrow \text{GatedMerge}(x_t, \{x_i\}_t)$ \COMMENT{\textcolor{gray}{$\triangleright$ Eqs. 4--9}}
    \ENDFOR
    \STATE $X[\text{targets}] \leftarrow x_t^{\text{new}}$; $\text{active} \leftarrow \text{active} \setminus \{\text{sources}\}$
    \STATE $E_{\text{sim}}[\text{targets}] \leftarrow \overline{X[\text{targets}] W_{\text{rand}}}$ \COMMENT{\textcolor{gray}{$\triangleright$ Refresh (Eq. 2)}}
\ENDWHILE
\RETURN $X_C \leftarrow X[\text{active}]$ with updated coordinates
\end{algorithmic}
\end{algorithm}

\subsection{Cross-Backbone Transfer Details}
\label{app:transfer}

CRAFT interacts with the host VLM through only two interfaces: (i) the visual token sequence after the vision encoder and multimodal projector, i.e., already in the language model's embedding space, and (ii) the spatio-temporal coordinates $(t,h,w)$ of each visual token. The compression core---the training-free selection of Algorithm~\ref{alg:lgm} and the learnable gated merging of \S\ref{sec:merging}---is implemented as a single backbone-agnostic module shared verbatim by all three backbones, so porting CRAFT to a new architecture only requires locating the visual span in the input sequence and supplying per-token coordinates. Because the weighting scorer and gate operate on the backbone's hidden width $D$, the lightweight merge module is re-instantiated at the target width and trained with exactly the same two-stage curriculum and hyperparameters as on Qwen3.5-4B (Table~\ref{tab:hparams}); no architecture-specific tuning is performed.

\noindent\textbf{Qwen2.5-VL-7B.} The visual token layout matches Qwen3.5 (patch embedding with $2{\times}2$ spatial merging, M-RoPE), so the port is direct: 3D position ids are obtained from the backbone's native \texttt{get\_rope\_index} (including Qwen2.5-VL's per-video temporal scaling), compression runs unchanged on the video span, and after merging the position ids are index-selected onto the surviving positions, so retained tokens keep their original 3D rotary coordinates.

\noindent\textbf{LLaVA-OV-7B.} This backbone differs in both the vision tower and the position encoding. Videos are encoded by SigLIP into a $27{\times}27$ grid per frame and pooled to $14{\times}14{=}196$ tokens per frame, followed by one \texttt{image\_newline} token per video; compression is applied to the $F{\times}196$ feature tokens only, and the newline token is always retained. The language model uses standard 1D RoPE rather than M-RoPE, so the $(t,h,w)$ coordinates required by the position-aware weighting scorer (Eq.~\ref{eq:qk}) are constructed explicitly from the known frame/grid layout and consumed only inside the scorer, while the language model keeps the original 1D positions: each surviving token inherits its own absolute position in the uncompressed sequence and decoding continues from the original sequence length, leaving all relative rotary distances identical to the uncompressed model.

That the same design and training recipe works across distinct vision encoders (ViT with $2{\times}2$ merging vs.\ SigLIP with pooling), token layouts, and position encodings (M-RoPE vs.\ 1D RoPE) substantiates the claim of \S\ref{sec:transfer} that CRAFT's effectiveness does not rely on architecture-specific tuning.

\section{Experiment Details}

\subsection{Benchmark Details}
\label{app:benchmarks}

\noindent\textbf{MVBench.} MVBench comprises $4{,}000$ QA pairs across $20$ temporal-understanding tasks. We evaluate on $3{,}800$ of them and skip the $200$ samples of the \texttt{episodic\_reasoning} task, which are provided as pre-extracted frame folders sampled at $\mathrm{fps}{=}3$ rather than as video files, inconsistent with the $\mathrm{fps}{=}2$ sampling used in our evaluation.

\noindent\textbf{TempCompass.} TempCompass offers several answer formats, including multiple-choice, yes/no, caption matching, and caption generation. We report on its full multiple-choice test set only.

\begin{figure*}[t]
    \centering
    \includegraphics[width=\textwidth]{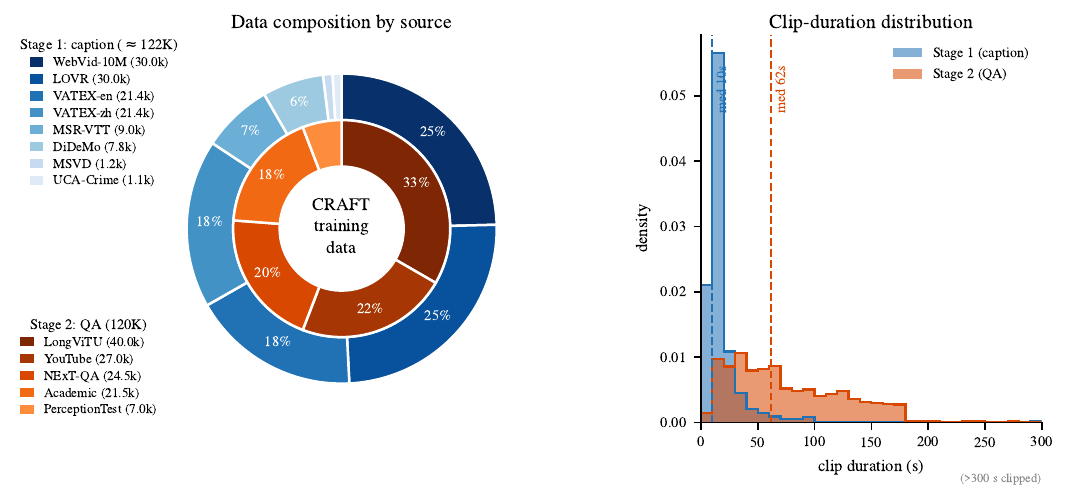}
    \caption{Training-data composition of the two-stage curriculum (\S\ref{sec:training}). \textbf{Left:} source composition as a nested ring---the outer ring (blues) is Stage-1 caption pre-training (${\approx}122$K video--caption pairs over eight public corpora), the inner ring (oranges) is Stage-2 QA (${120}$K pairs over five sources); percentages are within-stage shares and absolute counts are listed in the legend. \textbf{Right:} the Stage-2 temporal profile, resampled to a target long\,/\,medium\,/\,short mix of ${\approx}33\,/\,35\,/\,31\%$. Stage-2 sources play distinct roles: LongViTU (long-form), YouTube (open-domain breadth), NExT-QA (everyday), Academic (world knowledge), and PerceptionTest (fine-grained temporal).}
    \label{fig:datadist}
\end{figure*}

\subsection{Training Hyperparameters}
\label{app:hparams}

\begin{table}[t]\centering
\small
\setlength{\tabcolsep}{5pt}
\renewcommand{\arraystretch}{1.12}
\begin{tabular}{l c c}
\toprule
\textbf{Hyperparameter} & \textbf{Stage 1} & \textbf{Stage 2} \\
 & \textbf{Caption} & \textbf{QA} \\
\midrule
CRAFT LR      & $4\mathrm{e}{-}3$ & $6\mathrm{e}{-}6$ \\
LoRA LR (CRAFT$+$)       & $1\mathrm{e}{-}4$ & $1\mathrm{e}{-}4$ \\
Gate initialization      & from scratch & inherit Stage 1 \\
\midrule
Optimizer                & \multicolumn{2}{c}{AdamW} \\
LR schedule              & \multicolumn{2}{c}{cosine, warmup ratio $0.02$} \\
Weight decay             & \multicolumn{2}{c}{$1\mathrm{e}{-}2$} \\
Gradient clipping        & \multicolumn{2}{c}{$1.0$} \\
Epochs                   & \multicolumn{2}{c}{$1$} \\
Per-device batch size    & \multicolumn{2}{c}{$1$} \\
Gradient accumulation    & \multicolumn{2}{c}{$4$} \\
Global batch size        & \multicolumn{2}{c}{$32$ ($8{\times}$A800)} \\
Precision                & \multicolumn{2}{c}{bf16} \\
DeepSpeed                & \multicolumn{2}{c}{ZeRO-2} \\
Gradient checkpointing   & \multicolumn{2}{c}{enabled} \\
Sampling rate & \multicolumn{2}{c}{$\mathrm{fps}{=}2$} \\
Max frames & \multicolumn{2}{c}{$768$} \\
Max sequence length      & \multicolumn{2}{c}{$32$K} \\
\bottomrule
\end{tabular}
\caption{Training hyperparameters for the two-stage curriculum on Qwen3.5-4B. The upper block lists the settings that differ between stages; the lower block lists shared settings. Only the lightweight weighting scorer and gate are trained in both stages; CRAFT$+$ additionally trains LoRA adapters on the language model.}
\label{tab:hparams}
\end{table}

All experiments, including both training stages and all evaluations, are conducted on a single node with $8$ NVIDIA A800 (80GB) GPUs.
CRAFT is trained with a two-stage captioning-to-QA curriculum. Both stages fine-tune only the lightweight merge module---the position-aware weighting scorer and the content-adaptive gating network, a negligible fraction of the backbone parameters---inserted between the vision encoder and the language model, while the vision encoder, projector, and language-model backbone stay frozen throughout. We report two variants: \textbf{CRAFT}, which trains only this merge module, and \textbf{CRAFT$+$}, which additionally inserts LoRA adapters (rank $32$, $\alpha{=}64$, dropout $0.05$) into the language-model attention projections ($q,k,v,o$), trained with a separate learning rate. Stage~2 warm-starts from the Stage~1 checkpoint and inherits its learned gate, so it adapts the module with a much smaller learning rate. Table~\ref{tab:hparams} lists the full configuration.

During training the compression budget is not fixed. At every optimizer step we sample one compression setting and apply it consistently across the batch: with probability $0.5$ a similarity threshold drawn from $\{0.75,0.8,0.85,0.9,0.95\}$, and otherwise a fixed ratio drawn from $\{2,4,8,16,32\}$. This exposes the compressor to a wide range of budgets so that a single checkpoint generalizes across operating points.

\subsection{Training Data Distribution}
\label{app:data}

CRAFT's merge module is optimized with the two-stage curriculum of \S\ref{sec:training}: a caption pre-training stage followed by a video question-answering (QA) stage. Figure~\ref{fig:datadist} summarizes the composition of both stages; all data are drawn from public video corpora. To rule out data contamination, the training pool is audited against the six evaluation benchmarks and de-duplicated at the video level: videos overlapping any benchmark are removed, so the training set shares no videos---and hence no QA pairs---with the evaluation data; moreover, the compressed representation is query-agnostic, so no benchmark-specific data are used.

\noindent\textbf{Stage 1: caption pre-training.} We aggregate ${\approx}122$K video--caption pairs from eight public sources that span web video (WebVid-10M), bilingual described events (VATEX-en/zh, DiDeMo), short clips (MSR-VTT, MSVD), long-range video (LOVR), and surveillance (UCA-Crime). To keep the mixture broad and prevent high-volume corpora from swamping the long tail, each source is capped at $30$K pairs; WebVid-10M and LOVR are subsampled from much larger pools to this cap, while smaller sources are kept in full. This stage teaches the gate to preserve global semantics under aggressive merging.

\noindent\textbf{Stage 2: QA curriculum.} We then align the compressor on $120$K video-QA pairs drawn from five sources with complementary roles: world knowledge (Academic), open-domain breadth (YouTube), everyday activities (NExT-QA), long-form video (LongViTU), and fine-grained temporal perception (PerceptionTest); videos in Academic that overlap the existing NExT-QA/ActivityNet pools are de-duplicated. This mixture is deliberately shifted toward longer clips than Stage~1: measured durations rise from a Stage-1 median of $10$\,s ($96.5\%$ under $60$\,s) to a Stage-2 median of $62$\,s, with about half of the clips falling in the $1$--$3$\,min range (Fig.~\ref{fig:datadist}, right). This exposes the compressor to substantially more temporal content than the near-static short clips of Stage~1, steering it toward the fine-grained cues needed to answer questions.

\end{document}